\documentclass[11pt, a4paper, logo, onecolumn]{deepmind}
\usepackage{amsfonts}
\usepackage{mathtools}
\usepackage{wrapfig}
\usepackage{enumitem}

\usepackage{ntheorem}
\usepackage{dsfont}
\usepackage[dvipsnames]{xcolor}
\usepackage[colorinlistoftodos]{todonotes}
\usepackage{booktabs}
\usepackage{xfrac}
\usepackage{bbm}

\usepackage{algorithm}
\usepackage{algorithmicx}

\usepackage[most]{tcolorbox}

\newenvironment{WrapText}[1][r]
  {\wrapfigure{#1}{0.5\textwidth}\tcolorbox}
  {\endtcolorbox\endwrapfigure}

\usepackage[authoryear, sort&compress, round]{natbib} 
\title{Autocurricula and the Emergence of Innovation from Social Interaction:\\ A Manifesto for Multi-Agent Intelligence Research}
\correspondingauthor{jzl@google.com}

\paperurl{}

\reportnumber{} 

\renewcommand{\today}{March 2nd, 2019} 

\author[1]{Joel Z. Leibo}
\author[1]{Edward Hughes}
\author[1]{Marc Lanctot}
\author[1]{Thore Graepel}

\affil[1]{DeepMind}

\begin{abstract}
Evolution has produced a multi-scale mosaic of interacting adaptive units. Innovations arise when perturbations push parts of the system away from stable equilibria into new regimes where previously well-adapted solutions no longer work. Here we explore the hypothesis that multi-agent systems sometimes display intrinsic dynamics arising from competition and cooperation that provide a naturally emergent curriculum, which we term an autocurriculum. The solution of one social task often begets new social tasks, continually generating novel challenges, and thereby promoting innovation. Under certain conditions these challenges may become increasingly complex over time, demanding that agents accumulate ever more innovations.
\end{abstract}

\begin{document}
\maketitle
\balance



\section*{The Problem Problem}

\begin{WrapText}
{\large \textbf{Highlights}}
{\small
\begin{itemize}[leftmargin=*]
    \item General intelligence is connected to the ability to adapt and prosper in a wide range of environments.

    \item Generating new environments for research is labor-intensive and the current approach cannot scale indefinitely. Research progress is impeded by the ``problem problem’’.
    
    \item In social games, individuals must learn (a) which strategy to choose, and (b) how their strategy may be implemented by sequencing elementary actions.

    \item Ongoing strategic dynamics induce a sequence of implementation policy learning problems.
    
    \item The demands of competition and cooperation generate strategic dynamics.
\end{itemize}
}
\end{WrapText}

Pity the curious solipsist, for there is a limit to the knowledge they may acquire. To see why, consider a solitaire game played on a planar grid of size $19 \bigtimes 19$. The goal is to place black stones on the board so as to surround as much territory as possible. Obviously, the optimal solution is simply to place stones all along the edge of the grid. Imagine one learns by trial and error how to play this game. Once the optimal solution is discovered then there is nothing left to learn. The cleverness obtainable by practicing this game is bounded. Now introduce an additional player who places a white stone after each black stone. The white stones and the territories they enclose become barriers, preventing additional expansion of the black territory. Thus, the game of Go is born---a game with enough emergent complexity to occupy millions of minds for millenia \citep{fairbairn1995go}.

Intelligence may be defined as the ability to adapt to a diverse set of complex environments \citep{legg2007universal, hernandez2017measure}. This definition suggests that the ceiling of a solipsist's intelligence may only be raised by providing more and more environments of ever increasing diversity and complexity. To that end, recent work in artificial intelligence has relied on rich 3D simulation environments (e.g.~\cite{kempka2016vizdoom, beattie2016deepmind}). The resulting 
\begin{WrapText}
{\large \textbf{Glossary}}
{\small
\begin{itemize}[leftmargin=*]
    \item \textbf{Adaptive unit:} an umbrella term encompassing units of evolution and learning at any level of biological organization e.g., a species evolving genetically, a reinforcement learning agent, or a culturally evolving society.
    
    \item \textbf{Autocurriculum:} a self-generated sequence of challenges arising from the coupled adaptation dynamics of interacting adaptive units.
    
    \item \textbf{Challenge:} a change in the adaptive landscape faced by an adaptive unit.
    
    \item \textbf{Curriculum:} a sequence of challenges. Equivalently, a sequence of tasks chosen to direct learning.
    
    \item \textbf{Endogenous challenge:} a challenge arising from miscoordination or competition between an adaptive unit's component subunits.
    
    \item \textbf{Exogenous challenge:} a challenge arising from competition between adaptive units at the same hierarchical level.
    
    \item \textbf{Exploration by exploitation:} exploration that occurs as a byproduct of following the greedy policy estimate in a non-stationary environment.
    
    \item \textbf{Strategic choice:} a decision with game theoretic implications, e.g., to cooperate or defect. 
    
    \item \textbf{The problem problem:} the engineering problem of generating large numbers of interesting adaptive environments to support research.
    
    \item \textbf{Implementation policy:} a policy that implements a high-level strategic choice by sequencing elementary action primitives, e.g., movement actions.
    
    \item \textbf{Innovation:} an innovation expands an adaptive unit's behavioral repertoire with new robust and repeatable problem solving abilities.
    
    \item \textbf{Institution:} a system of rules, norms, or beliefs that determine the ``rules of the game'' played by the individuals composing a collective. The origination of a new institutions may be seen as a collective level innovation.
\end{itemize}
}
\end{WrapText}
agents have achieved proficiency at a wide range of tasks, such as navigating virtual mazes \citep{mirowski2016learning, savinov2018episodic}, foraging over rough, naturalistic terrain \citep{espeholt2018impala, hessel2018multi}, and tests drawn from the neuroscience and cognitive psychology literatures like visual search \citep{leibo2018psychlab} and memory recall \citep{wayne2018unsupervised}. As impressive as these results are, we think the research program they represent has fallen into a solipsistic trap. Just like the aforementioned Go-playing solipsist, the cleverness these agents may achieve is bounded by their adaptive environment(s). Advancing artificial intelligence by this route demands the creation of more and more environments, a laborious process similar to videogame design. Scaling up this process has become a bottleneck dubbed \textbf{the problem problem} (see Glossary).

How then did intelligence arise in nature? We propose that life solved its own version of the problem problem because it is a multi-agent system where any species's \textbf{innovation} determines the environment to which others must adapt. For example, it was early photosynthesizers that were the main source of atmospheric oxygen, setting the stage for the subsequent evolution of all the many organisms that depend on it for energy \citep{kasting2002life}. Likewise, human cultural evolution continually generates new ``rules of the game'', demanding continuous adaptation just to avoid being left behind by a changing world \citep{north1991institutions, gintis2000game, ostrom2005understanding, greif}.

The argument has two main ingredients. First, adaptive units must learn \textbf{implementation policies} for their high-level strategies by sequencing low-level action primitives \citep{leibo2017multiagent}. Second, the high level strategies themselves change over time in response to the \textbf{strategic choices} of others \citep{smith1973logic, schluter2000ecology, north1991institutions, gintis2000game}. Taken together, these two processes induce a sequence of challenges for the adaptive process that we term an \textbf{autocurriculum}. The rest of this paper is concerned with clarifying the autocurriculum concept and explaining how it provides a useful lens through which to view phenomena in evolutionary biology and multi-agent reinforcement learning. To that end, we offer a classification of the various kinds of autocurricula by their underlying social interaction (competition or cooperation). In the final part of the paper we consider the  conditions under which autocurricula might generate human-like accumulation of innovations.

\section*{Innovations Arise on All Levels of Life's Hierarchies}

\begin{figure}[t]
    \centering
    \includegraphics[width=1.0\textwidth]{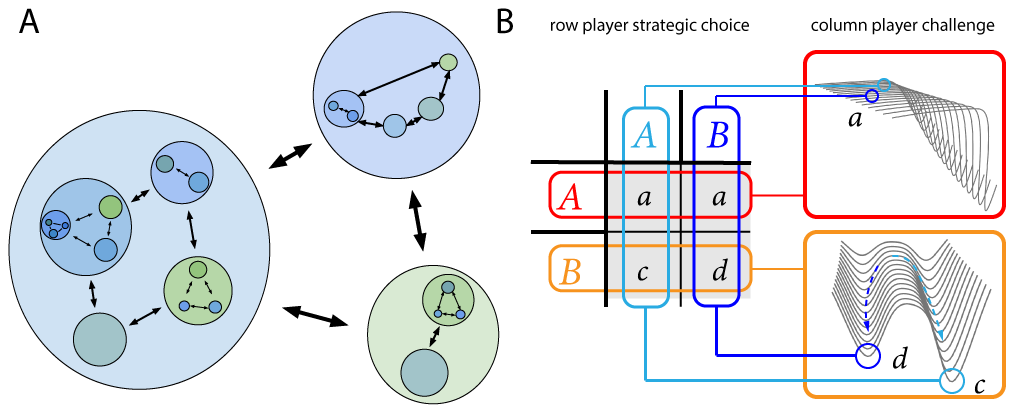}
\caption{(A) Adaptive units interacting with one another. Each adaptive unit is composed of sub-adaptive units and their interactions. Notice that this description is scale invariant. An adaptive subunit may be composed of interacting subsubunits which themselves may be composed by interactions of further subdivided units. (B) The relationship between row player strategic choices and column player implementation policy learning. When the row player shifts its strategy from $A$ to $B$, it induces a challenge for the column player. The optimal policy for the column player shifts to reflect its best response to $B$. In this case, an initially flat adaptive landscape where  outcome $a$ was achieved regardless of the column player's strategy, became a hilly landscape where the two strategies achieved different payoffs, $c$ for implementing $A$ and $d$ for implementing $B$.}
\label{fig:units_and_duality}
\end{figure}

Life on Earth is characterized by interactions between \textbf{adaptive units}. Each adaptive unit is composed of a set of interacting adaptive subunits, each of which is itself composed of interacting subsubunits and so on (Fig.~\ref{fig:units_and_duality}-A). For example, eukaryotic cells are composed of interacting prokaryotic organelles and human communities are made up of interacting individuals \citep{smith1997major, gerber1999holonic, ostrom2005understanding}. Consider the great feats of human intelligence---composing symphonies, sending astronauts to the moon, developing agricultural technology to feed billions of people. To which adaptive units should we attribute such success? One perspective, which we adopt here, postulates that these phenomena occur at the level of groups rather than individuals, i.e. human community, culture, or civilization. The truly intelligent adaptive units are massive multifaceted multi-agent systems that, while being composed of humans (as humans are composed of cells), have their own internal logic and computational mechanisms different from those of individual humans.

Innovations expand an adaptive unit's behavioral repertoire with new robust and repeatable problem solving abilities \citep{reader2003}, and may arise on any level of the hierarchy. A number of crucial innovations have had outsize influence on the subsequent history of life, for example the emergence of eukaryotic cells, multi-cellular organisms, and perhaps language \citep{smith1997major}. Human technological innovations like agriculture \citep{kavanagh2018hindcasting} and industry \citep{clark2008farewell} profoundly altered human lifestyles around the world. On the scale of human societies, the scope and effectiveness of \textbf{institutions} aimed at promoting cooperation have increased steadily over time \citep{pinker2011better}. The origins of such institutions like corporations, labor unions, parliamentary democracies, and inter-government organizations are all innovations of higher level adaptive units.

Some populations of adaptive units only change over time via relatively slow processes like genetic evolution (genes), while others adapt via much faster processes like reinforcement learning and cultural evolution (agents). Often interactions between processes on both timescales are significant, as in cases of gene-culture co-evolution \citep{boyd1988culture}. Many of the principles remain the same regardless of whether evolution or learning dominate \citep{borgers1997learning,  BloembergenTHK15, such2017deep}. We have adopted the unifying term ``adaptive unit'' to highlight the deep similarity between these processes, and to smoothly cross-apply insights originating in different fields.

Both evolution and reinforcement learning offer insights about how existing behaviors and solutions can be refined, for example by sharpening or stabilizing a desirable behavior. However, much less is known about how qualitatively novel innovations originate in the first place. The problem is that spontaneous generation of useful complex behavior is extraordinarily unlikely, yet necessary for the refinement processes of evolution and reinforcement learning to work their magic. Indeed the more complex the behavior, the lower the odds of its being generated spontaneously. This problem is exponentially exacerbated when two or more agents are required to jointly explore the set of solutions available to them as a group. Nevertheless, humans have walked on the moon, created the internet, and cured smallpox. But how?

\section*{Exploration by Exploitation}

Innovation may be explained by considering that the environment to which units adapt can change over time. This causes old adaptations to lose their shine and thereby motivates exploration toward new innovative solutions. Researcher-controlled non-stationary dynamics in machine learning, known as \textbf{curricula}, can facilitate the acquisition of complex behaviors that would not be learnable otherwise, e.g. \cite{asmuth2008potential, bengio2009curriculum, zaremba2015reinforcement, narvekar17, heess18}. The idea is to structure learning by changing the underlying environment over time.

We call such a change in the underlying environment dynamics a \textbf{challenge}. More precisely, we can think of a challenge to a policy as a change in its relative value compared to other policies. Notice that challenges may be positive as well as negative in nature. A previously successful predation strategy may diminish in effectiveness as prey species evolve countermeasures; or a chance dispersal to an uninhabited island may present an opportunity to apply foraging strategies that would not work on the mainland due to excessive competition.

Challenges motivate adaptive units to explore (and thus to learn) by following the gradient of their experience. That is, adaptive units explore because the true value of their current policy is shifting over time. We call this \textbf{exploration by exploitation}. In contrast to the traditional view in reinforcement learning based on the exploration-exploitation tradeoff, this view does not involve any deliberate trade-off between exploration and exploitation. An adaptive unit experiences new states not because it chooses to depart from exploitation, but because its underlying environment has changed.

Notice that a curriculum can be seen exactly as a sequence of challenges. We argue in this paper that certain kinds of curricula may emerge naturally from the non-stationary dynamics of social interaction processes, without any need for environmental engineering. We call such curricula autocurricula, since each challenge in the sequence is generated by the system itself. Adaptive social behavior of individuals continually reshapes the strategic landscape. We now seek to classify the various ways in which this can happen. The main distinguishing factor is whether the underlying challenge is endogenous or exogenous to the adaptive unit under consideration. This distinction underlies the contrasting and complementary dynamics of competition and cooperation.

\section*{Exogenous Challenges Generate Autocurricula}

An \textbf{exogenous challenge} is a challenge that originates outside the adaptive unit under consideration. For example, consider a two-player zero-sum game. Player one experiences an exogenous challenge when player two changes their strategy in such a way as to induce a change to player one's best response strategy. Viewed from the perspective of player one, this creates a change in its experienced adaptive landscape since its learning objective, implementing its best response strategy, has changed (Fig.~\ref{fig:units_and_duality}-B). Player two may shift strategy again once player one has successfully learned to implement its best response, thereby inducing yet another change in adaptive landscape (see Box 1). This interaction may produce a long sequence of novel challenges, an ``arms race'' \citep{dawkins1979arms} driving both players to improve their skills, i.e. an exogenous autocurriculum.

\begin{tcolorbox}
\textbf{Box 1. Duality of Strategy and Implementation\\}

The perspective developed here emphasizes situations where individuals must simultaneously learn not only what strategic choice to make, but also how to implement said choice by sequencing elementary actions. Thus, at any given time, the goal of each player's learning is to find an implementation for a strategy that best responds to the strategy profile of the others. Of course, since co-players do not just wait around for others to learn how they may be exploited, they too learn a best response. Whenever they change strategy they create a new adaptive landscape. Each such change constitutes a challenge. This creates a feedback loop. Any change has a cascading effect as others adjust their own strategy in response. Thus, from the perspective of an individual learner, the problem is one of adapting to a sequence of challenges, i.e. an autocurriulum. Innovation occurs when following an autocurriculum leads implementation learning to escape local optima where it would otherwise have been trapped indefinitely.\\

Empirical game theoretic techniques \citep{walsh2002analyzing, wellman2006methods} may be used to analyze behavior in terms of its strategic properties. Instead of assuming game rules are known a priori, these methods work backwards from outcome data to deduce properties of the game being played. For example, \cite{tuyls2018generalised} showed strategic intransitivities between AlphaGo versions and Zen, the previous state-of-the-art Go engine. A similar approach was taken for social dilemmas in \cite{leibo2017multiagent}. That work classified learned implementation policies by strategic properties such as ``aggressiveness''. \citep{perolat2017multi, hughes2018inequity} extended its approach beyond the two-player case to analyze the strategic incentives underlying policies learned by reinforcement learning in common pool resource appropriation and public goods scenarios.\\

Empirical game theoretic techniques are not only useful for data analysis, they also formed a critical part of one recent general-purpose algorithm for multi-agent reinforcement learning called Policy Space Response Oracles \citep{Lanctot17PSRO}. It works by incrementally building up the full normal form game table by iteratively adding to the table a best response to the mixed strategy equilibrium predicted for the table's previous state.
\end{tcolorbox}

However, there is no guarantee that novel challenges will continue to be generated in this way. Consider a game with intransitive strategic dynamics like rock-paper-scissors \citep{Singh00, Tuyls05Evolutionary}. The dynamics of evolving populations with incentives described by such games are often oscillatory (e.g.~\cite{gilpin1975limit}). That is, even though specific implementations may be different with each repetition, the same underlying challenges are continually repeated. This is an autocurriculum that endlessly chases its own tail, never breaking away to incentivize new innovations.

When do exogenous autocurricula break out of cycles and continually generate increasingly clever innovations? This question has been studied in multi-agent reinforcement learning in the framework of self-play algorithms for two-player zero-sum games. The idea behind this family of algorithms is that by continually training an adaptive unit to defeat past versions of itself it can always be paired with a partner of the appropriate difficulty, neither too strong nor too weak. Self-play ensures that the adaptive unit learns to exploit its own errors, thereby challenging itself to correct them the next time around. The algorithm TD-Gammon, which was the first to play competitively with human experts in the game of Backgammon, was an early example of this approach \citep{Tesauro95TDGammon}. Self-play remains a prominent approach in recent work. For example, \cite{bansal2018emergent} applied it to a 3D sumo wrestling game with simulated physics and \cite{jaderberg2018human} applied it to an egocentrically-viewed team Capture-the-flag game based on the Quake game engine.

However, in designing self-play algorithms, care must be taken to prevent forgetting of past policies. If old policies are forgotten, then a newer generation may become unable to defeat an older generation, creating an opening for long-extinct traits to re-appear in the future \citep{Samothrakis13}. Thus forgetting may cause the induced autocurriculum to double back onto itself, just like it does in the rock-paper-scissors case, preventing the productive accumulation of new innovations. In practice, successful self-play algorithms generally play not just against the latest (and strongest) policy, but also against as large and diverse as possible a set of older policies \citep{Lanctot17PSRO}.

In games with a small number of possible  strategies, an exogenous autocurriculum is expected to approach a Nash equilibrium \citep{gintis2000game}. However, in more open games, or those with a huge but still finite space of possible strategies, like Go or Chess, then self-play really does seem to be able to continually generate novel innovations. For example, AlphaGo and its Chess/Shogi/Go-playing variant, AlphaZero \citep{Silver16Go,silver2017agz,silver2018general} are based on self-play. Starting with adaptive policy and value estimators for game positions, they use Monte Carlo tree search to improve the current policy, then learn better estimators from the generated games. Interestingly, these algorithms show that some kinds of forgetting are not always harmful. Sometimes innovations are discovered at one point in the training process, and later on discarded in favor of others. AlphaGo Zero, for example, rediscovered several patterns known from human Go expertise called joseki, but some of them were discarded in favour of new variations later on in training \citep{silver2017agz}. A similar phenomenon was observed in AlphaZero: its preferences towards certain Chess openings fluctuated in time; the most frequently played openings at certain intermediate points in training were no longer seen in its later stages \citep{silver2018general}. Presumably the algorithm discovered that the discarded strategies are suboptimal. Sometimes it goes beyond our human capacity to understand what it has found. For instance, AlphaZero, in Chess, makes surprising (to humans) sacrifices to gain positional advantage, a development that is now impacting human professional level Chess \citep{kasparov2018chess}. 

So far we have only discussed exogenous autocurricula in the context of two-player zero sum games. A recent paper describing an algorithm called Malthusian reinforcement learning considered them in more general settings \citep{leibo2018malthusian}. Malthusian reinforcement learning extends self-play to allow for variable numbers of players to appear in each episode. Subpopulation sizes grow proportionally to their success. In games with limited resources this demographic expansion creates additional competitive pressure. That is, it induces an exogenous autocurriculum. Whenever a successful policy arises at any  population size, its own success ensures that population will increase in the future. Thus the Malthusian reinforcement learning algorithm generates a continually changing adaptive landscape which functions to perturb old solutions that have grown too comfortable and thereby driving it to escape poor local optima where state-of-the-art single-agent methods cannot avoid becoming stuck. 

Exogenous autocurricula also appear in some evolutionary models of human intelligence. In one theory, the main selection pressure on intelligence in human evolution is posited to be the need to manipulate others within the social group in order to climb a dominance hierarchy \citep{humphrey1976social}. Increasingly clever social partners motivate the need to evolve still greater manipulating intelligence, and so on, increasing intelligence up to the point where brain size could no longer expand for anatomical reasons \citep{byrne1996machiavellian, dunbar2017there}. On the other hand, there is more to being human than competition. As we will see in the next section, the challenges of organizing collective action also yield autocurricula that may have structured the evolution of human cognitive abilities and motivated significant innovations. 

\section*{Endogenous Challenges Generate Autocurricula}

Autocurricula may emerge on any level of the hierarchy of adaptive units. When a level is atomic (indivisible), only exogenous challenges are possible. On all other levels, the adaptive units are made up of adaptive subunits. In such cases adaptation may also be driven in response to \textbf{endogenous challenges} to the collective's integrity. For example, a collective-level adaptive unit will generally function best when it has suppressed most competition between its component subunits. In multi-cellular organisms, that suppression sometimes breaks down, freeing somatic cells to behave in their own short-sighted selfish interest \citep{frank2004problems, rankin2007tragedy}. Cancerous cells often behave like unicellular organisms, even reverting to less efficient fermentation-based metabolism (the Warburg effect) \citep{vander2009understanding} and activating other ancient cellular functions conserved in modern unicellular organisms \citep{trigos2018evolution}. Similar breakdowns of cooperation can occur on the level of a society. For example, eusocial insect colonies are vulnerable to exploitation by renegade worker reproduction \citep{beekman2008workers}.

Such situations are social dilemmas. They expose tensions between individual and collective rationality \citep{rapoport1974prisoner}. One particularly well-studied type of social dilemma is called common-pool resource appropriation \citep{ostrom1990governing}. For a common-pool resource like a common grazing pasture, fishery, or irrigation system, it is difficult or impossible for individuals to exclude one another's access. But whenever an individual obtains a benefit from such a common-pool resource, the remaining amount available for appropriation by others is at least somewhat diminished. If each individual's marginal benefit of appropriation exceeds their share of the cost of further depletion, then they are predicted to continue their appropriation until the resource becomes degraded. This situation is called the tragedy of the commons \citep{hardin1968tragedy, ostrom1990governing}. It is impossible for an individual acting unilaterally to escape this fate; since even if one were to restrain their appropriation, the effect would be too small to make a difference. Thus individual-level innovation is not sufficient to meet this challenge. Any innovation that resolves a social dilemma must involve changing the behavior of a critical fraction of the participants \citep{schelling1973hockey}. 

One way to effect such a change in the joint behavior of many individuals is to originate an ``institution'' \citep{ostrom1990governing, ostrom2005understanding, greif}: a system of rules, norms, or beliefs. Institutions may structure individual-level adaptive processes to ensure the group as a whole achieves a socially beneficial outcome. They may be regarded as collective-level innovations. For example, consider an institution whereby individuals who over-exploit the common pool resource are sanctioned by the group. This institution changes the individual incentive structure such that over-exploiting is no longer the dominant strategy. We have seen hints of emergent institutions in recent multi-agent reinforcement learning models of common-pool resource appropriation situations \citep{leibo2017multiagent, perolat2017multi, hughes2018inequity}. For example. an institution for generating cooperation on the collective-level emerged in \cite{hughes2018inequity} when certain agents learned to sanction over-exploiters, effectively policing their behavior.

\begin{tcolorbox}
\textbf{Box 2. No-Free-Lunch in Social Dilemmas\\}

The literature contains several models that suggest that higher order social dilemmas can be evaded in various ways. Here we show that they depend on unrealistic assumptions, thereby sustaining the present argument that the no-free-lunch property of social dilemmas cannot generally be avoided.\\

For example, some models depend on asymmetries between altruistic cooperation and altruistic punishment \citep{boyd2003evolution}. This may hold in some situations, especially when the cost of monitoring for infractions is small, e.g. in agricultural land use. But it does not hold when monitoring costs are large, as in many real-world common-pool resource situations \citep{ostrom1990governing}.\\

Other models depend on coordination of large numbers of individuals engaging in altruistic punishment \citep{boyd2010coordinated}, but they do not take into account the costs of such coordination. Several of the case studies of effective community-level resource management described by Ostrom show that communities are willing to invest in complex and relatively costly institutional mechanisms for ensuring this coordination is effective so that sanctioning may be deemed legitimate by the group and no individual must bear the brunt of the cost \citep{ostrom1990governing}.\\ 

Another intriguing idea is to link reputation across tasks \citep{panchanathan2004indirect}. However, these mechanisms substantially increase pressure on institutions for assigning and communicating reputations. Thus they give rise to new attack vectors for the unscrupulous. Agents may try to cheat by finding ways to falsely inflate their reputations.\\

While these arguments may explain some instances of cooperation, especially when the costs of monitoring for infractions are low, they are insufficient to explain away the no-free-lunch principle that generates higher order social dilemmas.
\end{tcolorbox}

A social dilemma may be resolved via the emergence of an institution that systematically changes payoffs experienced by adaptive units so as to eliminate socially deficient equilibria or nudge learning dynamics toward better equilibria. However, maintaining the institution itself still depends on interactions of those same participants. In many cases this yields a second order social dilemma because each individual would prefer others to shoulder a greater share of that burden \citep{axelrod1986evolutionary, yamagishi1988seriousness, heckathorn1989collective}. This is called the ``second-order free rider problem''. As predicted by these models, there is evidence that pre-state societies sustain social norms that disapprove of second-order free riding, e.g.~\cite{mathew2017second}. Second-order social dilemmas may themselves be resolved via the emergence of higher order institutions which, in turn, create their own still higher level successor dilemmas \citep{ostrom2000collective}. Indeed we can say that social dilemma situations have a kind of ``no-free lunch'' property: once you resolve a social dilemma in one place then another one crops up somewhere else (see Box 2). A society may resolve a social dilemma by hiring watchmen, but then who watches the watchmen? These dynamics may generate a sequence of endogenous challenges growing steadily in scope and complexity, i.e. an autocurriculum.

Just as atomic individuals participate in social interactions with one another, communities interact with peer communities as competitors or allies. Exogenous challenges may also arise on the collective level. Their effects reverberate down the hierarchy, tending to resolve endogenous challenges by aligning the incentives of lower level entities \citep{smith1997major, henrich2004, wilson2007rethinking}. For example, eusocial insect colonies compete with other conspecific colonies. Those colonies that are better able to maintain internal cooperation, e.g., by establishing reproductive division of labor, are more likely to be successful in colony-level competition \citep{nowak2010evolution}. In this view, communities are treated much like atomic individuals. Just as atomic individuals are understood to seek to maximize the utility (food, shelter, mates, etc) they can obtain from social interaction, higher order adaptive units act to optimize a range of different higher order utility concepts. For example, populations of ant colonies optimize fitness in the sense of multi-level selection \citep{okasha2005multilevel}, while human corporations optimize culturally-determined constructs like ``shareholder value''. Analogous to nervous systems, communities weigh their options via decision-making institutions like parliaments and markets.

\section*{Accumulating Autocurricula and Human Uniqueness}
\begin{WrapText}
{\large \textbf{Outstanding Questions}}
{\small
\begin{itemize}[leftmargin=*]
    \item Can autocurricula generate sufficiently diverse challenges to resolve the problem problem?

    \item Does the duality between strategy and implementation persist at the level of the community?

    \item Can the no-free-lunch property of social dilemmas be formalized? What new experiments could be carried out to demonstrate or refute its validity?
    
    \item Did autocurricula phenomena play a role in the evolution of higher-order individuals like multi-cellular organisms and eusocial insect societies? Could analogous  transitions arise in multi-agent reinforcement learning?
    
    \item How do challenges arising on different levels of the biological hierarchy interact with one another? For example, do higher order exogenous challenges align the interests of lower order individuals? What happens if lower order individuals can defect from their ``team'' to join another, more successful, higher order individual?
    
    \item How can we establish feedback loops like cumulative culture and self-domestication in silico?
\end{itemize}
}
\end{WrapText}
In this paper we argued that interactions among multiple adaptive units across levels of the biological hierarchy give rise to sequences of challenges called autocurricula that perturb adaptive landscapes. This enables adaptive units to discover innovations by continually adapting to changing circumstances---exploration-by-exploitation (see Outstanding Questions). Might this mechanism be the key to solving the problem problem? Will autocurricula generate enough adaptive diversity? Perhaps not. Recall that autocurricula may be cyclic, repeatedly learning and unlearning the same information, innovations never accumulating or building on one another. Moreover, there is nothing about these mechanisms that suggests they do not apply equally strongly to non-human animals. We think the solution is as follows. Autocurricula do indeed exist throughout animal evolution. However, humans are unique among animals in their exceptionally long cultural memory. This allows intransitive cycles to be avoided, thereby promoting cumulative accumulation of innovation after innovation.

Why did this same accumulation not occur in other ape species? We highlight two possibilities, based on the structure of feedback loops within human societies, driven respectively by exogenous and endogenous challenges to group integrity. Both may be seen as the combination of a challenge-based loop, and a ratchet loop that serves the purpose of accumulating and distilling beneficial innovations.

\begin{figure}[H]
    \centering
    \includegraphics[width=0.95\textwidth]{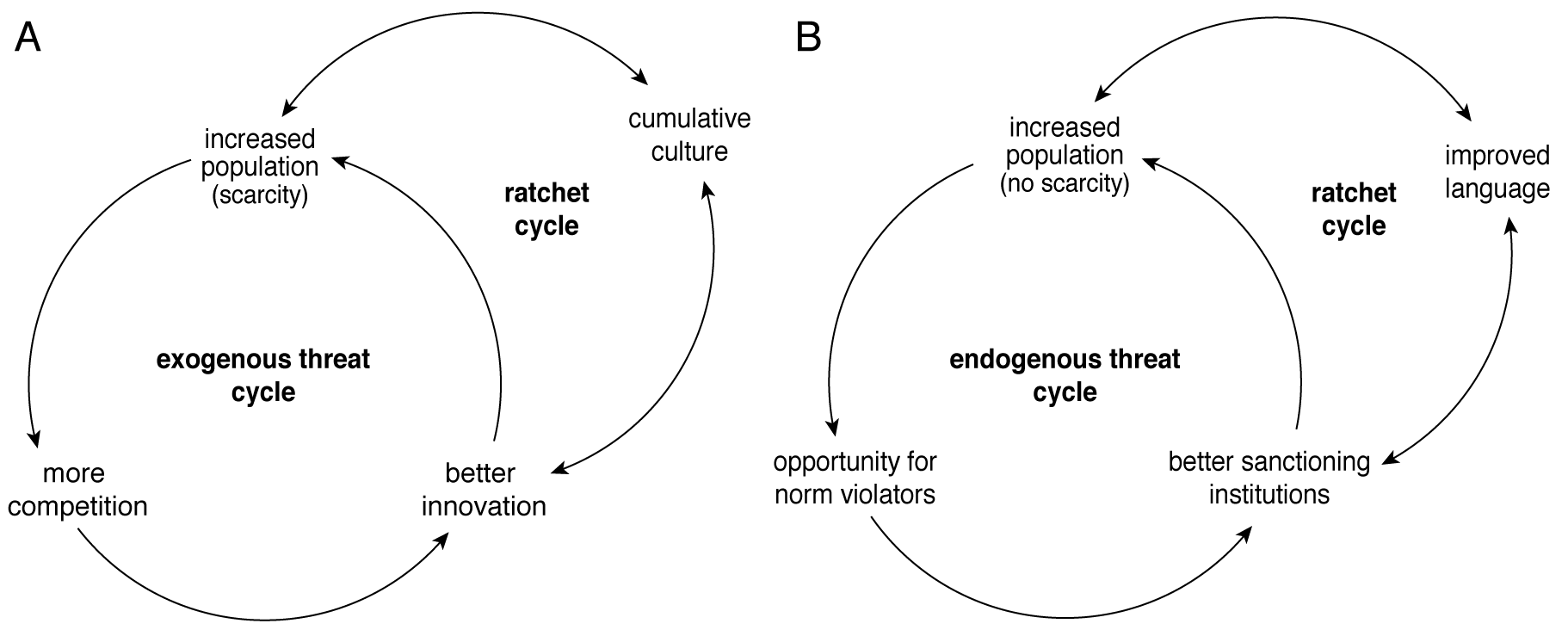}
\caption{(A) The cumulative culture loop. (B) The self-domestication loop.}
\label{fig:feedback_loops}
\end{figure} 

\begin{enumerate}
    \item \textbf{The cumulative culture loop (Fig. \ref{fig:feedback_loops}-A).} A growing population puts stress on human societies since more individuals must vie for access to the same limited resources. The resulting competitive pressure may give rise to exogenous challenges to each individual, motivating new innovation. However, since humans are adept social learners, any innovation that promotes survival will tend to quickly spread through the population and thereby increase population sizes in the future along the lines suggested by models like \cite{enquist2011modelling}. There is a significant literature demonstrating that possibly uniquely human high fidelity social transmission of cultural elements allows innovations to build upon one another and increase in complexity over time \citep{boyd2011cultural, dean2012identification, boyd2013facts, muthukrishna2016innovation}. Moreover, archaeological evidence and computational models suggest that larger populations can afford greater protection from random forgetting of cultural elements due to chance events like the death of a single master artisan, thus allowing a more significant opportunity for cumulative cultural evolution \citep{henrich2004demography}. Indeed, cultural evolution is made more effective by increasing the population size and increasing the population size increases the effectiveness of cultural learning. This is a feedback loop that, once started, could potentially increase until the population becomes limited by its environment's carrying capacity in some way that further innovation cannot transcend.
    
    \item \textbf{The self-domestication loop (Fig. \ref{fig:feedback_loops}-B).} A growing population need not lead to scarcity, provided that resource production is efficient. However, it does provide more opportunity for norm violators to free-ride on the contributions of others \citep{carpenter2007}. One norm-violating behavior is reactive aggression, particularly in young males. Coordinated punishment for aggression in small-scale societies often takes the form of capital punishment \citep{boehm2011retaliatory}. The result is genetic evolution that gradually reduces the prevalence of aggressive individuals. This process has been called self-domestication \citep{wrangham2018two} because it is similar to the selection process that may have been applied in creating domestic dogs from wolves, now applied by a species to itself. Self-domestication has the effect of increasing tolerance for non-kin because it reduces the likelihood of encountering aggressive individuals. This may have created opportunities for improving communication abilities, and ultimately, language. Conversely, improving communication between individuals in a society improves the effectiveness of institutions for sanctioning norm violators by facilitating coordinated punishment \citep{bochet2006communication, janssen2010lab, janssen2014effect}. Language also improves the accuracy of reputation information, e.g. as conveyed by gossip, informing the decision of which individuals deserve punishment \citep{nowak1998evolution, rockenbach2006efficient}. Hence, the resultant feedback loop may accumulate institutions indefinitely.
\end{enumerate}

In this paper we identified multi-agent interactions as key drivers of sustained innovation and, perhaps, increases in  intelligence over the course of human evolution. As a consequence, a research program based only on replicating individual human cognitive abilities, e.g. attention, memory, planning, etc, is likely incomplete. It seems that intelligence researchers would do well to pay more attention to the ways in which multi-agent dynamics may structure both evolution and learning.

\section*{Acknowledgements}

The authors would like to thank Edgar Duenez-Guzman, David Balduzzi, Aliya Amirova, Greg Wayne, Peter Sunehag, Joyce Xu, Martin Chadwick, Richard Everett, Vlad Firiou and Raphael Koster for very helpful comments and discussions during the drafting of this article. In addition, the first author would like to thank all the speakers, organizers, and attendees of the 2014 ``Are there limits to evolution?'' workshop at Cambridge where much of the thought process leading eventually to this article was first hatched.

\bibliography{main}

\end{document}